\documentclass[10pt,logo,copyright]{giga-report}
\usepackage[authoryear,sort&compress,round]{natbib}

\usepackage[utf8]{inputenc} %
\usepackage[T1]{fontenc}    %

\usepackage{parskip}        %
\usepackage{url}            %
\usepackage{booktabs}       %
\usepackage{amsfonts}       %
\usepackage{nicefrac}       %
\usepackage{microtype}      %
\usepackage{xcolor}         %
\usepackage[dvipsnames]{xcolor} %
\usepackage{graphicx}
\usepackage{animate}        %
\usepackage{subcaption}
\usepackage{tabularx}
\usepackage{makecell}
\usepackage{adjustbox}
\usepackage{setspace}
\usepackage{todonotes}
\usepackage{colortbl}
\usepackage{wrapfig}

\newcolumntype{M}[1]{>{\centering\arraybackslash}m{#1}}
\usepackage{float}
\usepackage{tikz}
\usetikzlibrary{positioning,shapes,arrows}
\usepackage{amsmath,amsfonts,bm, bbm,leftindex}
\usepackage{multirow}
\usepackage{comment}
\usepackage{gensymb}
\usepackage{lipsum}
\usetikzlibrary{arrows.meta, positioning, fit}
\usepackage[para]{threeparttable}
\usepackage{tikz}
\usetikzlibrary{tikzmark}

\def\eqref#1{equation~\ref{#1}}

\def\1{\bm{1}}

\DeclareMathAlphabet{\mathsfit}{\encodingdefault}{\sfdefault}{m}{sl}
\SetMathAlphabet{\mathsfit}{bold}{\encodingdefault}{\sfdefault}{bx}{n}

\makeatletter
\let\save@mathaccent\mathaccent
\newcommand*\if@single[3]{%
  \setbox0\hbox{${\mathaccent"0362{#1}}^H$}%
  \setbox2\hbox{${\mathaccent"0362{\kern0pt#1}}^H$}%
  \ifdim\ht0=\ht2 #3\else #2\fi
  }
\newcommand*\rel@kern[1]{\kern#1\dimexpr\macc@kerna}
\newcommand*\widebar[1]{\@ifnextchar^{{\wide@bar{#1}{0}}}{\wide@bar{#1}{1}}}
\newcommand*\wide@bar[2]{\if@single{#1}{\wide@bar@{#1}{#2}{1}}{\wide@bar@{#1}{#2}{2}}}
\newcommand*\wide@bar@[3]{%
  \begingroup
  \def\mathaccent##1##2{%
    \let\mathaccent\save@mathaccent
    \if#32 \let\macc@nucleus\first@char \fi
    \setbox\z@\hbox{$\macc@style{\macc@nucleus}_{}$}%
    \setbox\tw@\hbox{$\macc@style{\macc@nucleus}{}_{}$}%
    \dimen@\wd\tw@
    \advance\dimen@-\wd\z@
    \divide\dimen@ 3
    \@tempdima\wd\tw@
    \advance\@tempdima-\scriptspace
    \divide\@tempdima 10
    \advance\dimen@-\@tempdima
    \ifdim\dimen@>\z@ \dimen@0pt\fi
    \rel@kern{0.6}\kern-\dimen@
    \if#31
      \overline{\rel@kern{-0.6}\kern\dimen@\macc@nucleus\rel@kern{0.4}\kern\dimen@}%
      \advance\dimen@0.4\dimexpr\macc@kerna
      \let\final@kern#2%
      \ifdim\dimen@<\z@ \let\final@kern1\fi
      \if\final@kern1 \kern-\dimen@\fi
    \else
      \overline{\rel@kern{-0.6}\kern\dimen@#1}%
    \fi
  }%
  \macc@depth\@ne
  \let\math@bgroup\@empty \let\math@egroup\macc@set@skewchar
  \mathsurround\z@ \frozen@everymath{\mathgroup\macc@group\relax}%
  \macc@set@skewchar\relax
  \let\mathaccentV\macc@nested@a
  \if#31
    \macc@nested@a\relax111{#1}%
  \else
    \def\gobble@till@marker##1\endmarker{}%
    \futurelet\first@char\gobble@till@marker#1\endmarker
    \ifcat\noexpand\first@char A\else
      \def\first@char{}%
    \fi
    \macc@nested@a\relax111{\first@char}%
  \fi
  \endgroup
}
\makeatother

\definecolor{darkred}{rgb}{0.7, 0.0, 0.0}

\usepackage{amsmath} 

\usepackage[table]{xcolor}
\newcommand{\cmark}{\textcolor[HTML]{003399}{\ding{51}}}
\newcommand{\xmark}{\textcolor{gray!70!black}{\ding{55}}}
\usepackage{pifont}
\usepackage{graphicx}
\newcommand{\rot}[1]{{\scriptsize #1}}

\usepackage[nameinlink]{cleveref}
\crefname{equation}{Eq.}{Eqs.}
\crefname{figure}{Fig.}{Figs.}
\crefname{section}{Sec.}{Sec.}
\crefname{appendix}{App.}{App.}
\crefname{table}{Tab.}{Tabs.}
\crefname{algorithm}{Algo}{Algo}
\crefname{thm}{Thm}{Thm}
\Crefname{thm}{Thm}{Thm}
\crefname{prop}{Prop}{Prop}

\newcommand{\crefnames}[3]{%
  \@for\next:=#1\do{%
    \expandafter\crefname\expandafter{\next}{#2}{#3}%
  }%
}

\title{GigaBrain-0: A World Model-Powered Vision-Language- Action Model}

\author{
\vspace{-0.1in}
\centerline{GigaAI} 
\centerline{{Project Page: \href{https://gigabrain0.github.io}{https://gigabrain0.github.io}}} 
\footnotesize
\textbf{GigaBrain Team (alphabetical order)}:
\normalfont
Angen Ye,  
Boyuan Wang,  
Chaojun Ni,  
Guan Huang,  
Guosheng Zhao,  
Haoyun Li,  
Jie Li,  
Jiagang Zhu,  
Lv Feng,  
Peng Li,  
Qiuping Deng,  
Runqi Ouyang,  
Wenkang Qin,  
Xinze Chen,  
Xiaofeng Wang,  
Yang Wang,  
Yifan Li,  
Yilong Li,  
Yiran Ding,  
Yuan Xu,  
Yun Ye,  
Yukun Zhou,  
Zhehao Dong,  
Zhenan Wang,  
Zhichao Liu,
Zheng Zhu.
\vspace{-1em}
}

\begin{document}
\maketitle

\begin{center}
    \centering
    \captionsetup{type=figure, justification=justified, singlelinecheck=false}
    \includegraphics[width=0.95\linewidth]{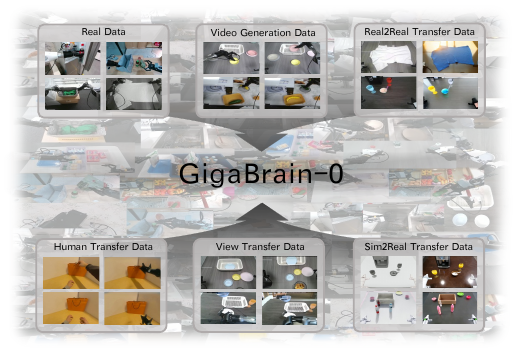}
\vspace{-0.2in}
    \caption{\textit{GigaBrain-0} is a Vision-Language-Action (VLA) model trained on real-world robot data and diverse data generated by world models, including video generation data, Real2Real transfer data, human transfer data, view transfer data, and Sim2Real transfer data, to enhance its generalization in real-world environments.}
    \label{fig:teaser}
\end{center}

\begin{abstract}
\vspace{-0.1in}
Training Vision-Language-Action (VLA) models for generalist robots typically requires large-scale real-world robot data, which is expensive and time-consuming to collect. The inefficiency of data collection severely limits the scalability, and generalization capacity of current VLA systems. Therefore, we introduce \textit{GigaBrain-0}, a novel VLA foundation model empowered by world model-generated data. By leveraging world models to generate diverse data at scale, \textit{GigaBrain-0} significantly reduces reliance on real robot data while improving cross-task generalization.
Our approach further improves policy robustness through RGBD input modeling and embodied Chain-of-Thought (CoT) supervision, enabling the model to reason about spatial geometry, object states, and long-horizon dependencies during task execution. This leads to substantial gains in real-world performance on dexterous, long-horizon, and mobile manipulation tasks. Extensive experiments demonstrate that \textit{GigaBrain-0} achieves 
superior generalization across variations in appearances (e.g., textures, colors), object placements, and camera viewpoints. Additionally, we present \textit{GigaBrain-0-Small}, an optimized lightweight variant designed to run efficiently on devices such as the NVIDIA Jetson AGX Orin.
\end{abstract}

\abscontent
\section{Introduction}
Recent advances in vision-language-action (VLA) models~\citep{pi0,pi05,go1,gr00t,gr3,walloss,galaxea} have shown promising results in enabling generalist robots to understand high-level instructions, perceive their environments, and execute complex manipulation tasks. These models aim to bridge the gap between symbolic reasoning and embodied action by integrating visual inputs, natural language commands, and motor control into a unified framework. However, training such models typically relies on large-scale datasets of real-world robot interactions, and such data collection is not only expensive and time-consuming but also limited in diversity and scalability. The inefficiency of physical data collection poses a fundamental bottleneck to the development of robust, general-purpose robotic systems capable of operating across a wide range of environments, objects, and task configurations.

To overcome these limitations, we introduce \textit{GigaBrain-0}, a novel VLA foundation model that leverages world model-generated data to reduce reliance on costly real-world robot data while improving generalization and data efficiency. By training on synthetic yet realistic trajectories generated by world models, as shown in Fig.~1, \textit{GigaBrain-0} accesses a vast and diverse set of experiences, including variations in object materials, colors, lighting, and viewpoints, far beyond what is feasible to collect physically. This scalable data generation pipeline enables the model to learn robust representations that transfer effectively to real-world environments.

Our approach further enhances policy robustness through two framework innovations: RGBD input modeling and embodied Chain-of-Thought (CoT) supervision. By incorporating depth information, the model gains a richer understanding of 3D geometry and spatial layout, which is crucial for precise manipulation. Meanwhile, the embodied CoT framework encourages the model to generate intermediate reasoning steps, such as manipulation trajectories and subgoal planning, mimicking the cognitive processes underlying human problem-solving. This structured reasoning enables effective handling of long-horizon tasks and fine-grained actions that require sustained attention and sequential decision-making.

We evaluate \textit{GigaBrain-0} through extensive real-world robotic deployments, including dexterous manipulation tasks (e.g., laundry folding, paper towel preparation), long-horizon tasks (e.g., table bussing, juice preparation), mobile manipulation tasks (e.g., boxes moving, laundry baskets moving). 
Results show that \textit{GigaBrain-0} delivers consistently strong performance across this broad range of tasks and exhibits exceptional generalization under diverse conditions, including changes in appearances (e.g., texture, color), object placements, and camera viewpoints. Furthermore, we introduce \textit{GigaBrain-0-Small}, an efficient variant optimized for deployment on hardware like the NVIDIA Jetson AGX Orin. 
Our work highlights the potential of world model-generated data as a scalable and effective alternative to traditional data collection paradigms, marking a significant step toward versatile, general-purpose robotic systems.

\section{Related Works}

\subsection{Vision-Language-Action Models}
Developing general-purpose robotic manipulation policies that can interpret high-level instructions and operate effectively in diverse physical environments remains a fundamental challenge in robotics. Prior approaches~\citep{r3m, radford2021learning, xiao2022masked, karamcheti2023language} have focused on leveraging heterogeneous datasets to learn rich, transferable representations that enhance policy robustness in complex and unseen scenarios. Inspired by advances in large language models, VLA frameworks~\citep{o2024open, team2024octo, kimopenvla, pi0, pertsch2025fast, pi05, doshi2024scaling, wang2024scaling, liu2024rdt, qu2025spatialvla, li2024cogact,gr00t,gr3} have emerged as a promising paradigm, scaling model capacity and data volume to improve generalization across tasks and embodiments. These models typically build upon pretrained vision-language models \citep{qwen25vl,paligemma,kosmos,llava,flamingo}, integrating multimodal inputs to generate action sequences, either through autoregressive token prediction or continuous action modeling via flow matching \citep{lipman2022flow,liu2022rectified}. This shift from purely vision-to-action architectures to semantically grounded, language-conditioned policies has significantly expanded the scope of achievable behaviors. To support such data-hungry models, researchers have increasingly relied on cross-embodiment datasets ~\citep{o2024open, khazatsky2024droid, dasari2019robonet, walke2023bridgedata, ebert2021bridge}, combining publicly available logs from heterogeneous robotic platforms as well as large proprietary collections, such as those used in \citep{gr3,pi0,pi05,gr00t}. Despite the performance gains, this reliance on extensive real-world interaction data raises practical concerns regarding scalability and cost.

\subsection{World Models as Data Engines}
The recent advancement in world model development~\citep{hunyuanvideo,wan,cosmos,cosmospredict,vjepa2,genieEnvisioner,enerverseac,dreamgen} has catalyzed a shift toward using synthetic data to narrow the sim-to-real divide in embodied intelligence~\citep{sorasurvey}. In fields like autonomous driving, generative models are increasingly employed to simulate complex traffic scenarios, as evidenced by approaches such as~\citep{drivedreamer,drivedreamer2,gaia,gaia2,magicdrive,vista,drivedreamer4d,recondreamer,recondreamer++,cosmosdrivedream,recondreamerrl}. In robotics, where data acquisition is often limited by hardware and labor intensity, generative world models offer a promising alternative. Works including~\citep{unisim,unipi,robodreamer,vidar,dreamgen} leverage natural language prompts to generate plausible future trajectories, which are then converted into low-level control signals via inverse dynamics estimation. To improve the structural fidelity of synthesized outputs, TesserAct~\citep{tesseract} and Robot4DGen~\citep{Robot4DGen} introduce a unified multimodal generation framework that produces synchronized streams of RGB frames, depth maps, surface normals, or 3D point clouds. This enables the construction of temporally coherent 4D reconstructions, significantly advancing policy learning compared to models trained on RGB video alone. Further enhancements in scene diversity are achieved through techniques like background inpainting~\citep{rebot,roboengine}, which alters environmental textures, and video-to-video translation methods~\citep{robotransfer,embodiedreamer,cosmos,mimicdreamer,egodemogen,emma} that stylize or adapt visual dynamics. Notably, \textit{GigaBrain-0} exploits the generative capacity of world models to produce highly varied data across texture, material, illumination, object placement, and camera viewpoints, providing rich, generalizable data source for training VLA models.
\section{GigaBrain-0 Model}

\begin{figure}[!t]
    \centering
    \vspace{-0.5cm}
    \includegraphics[width=1\linewidth]{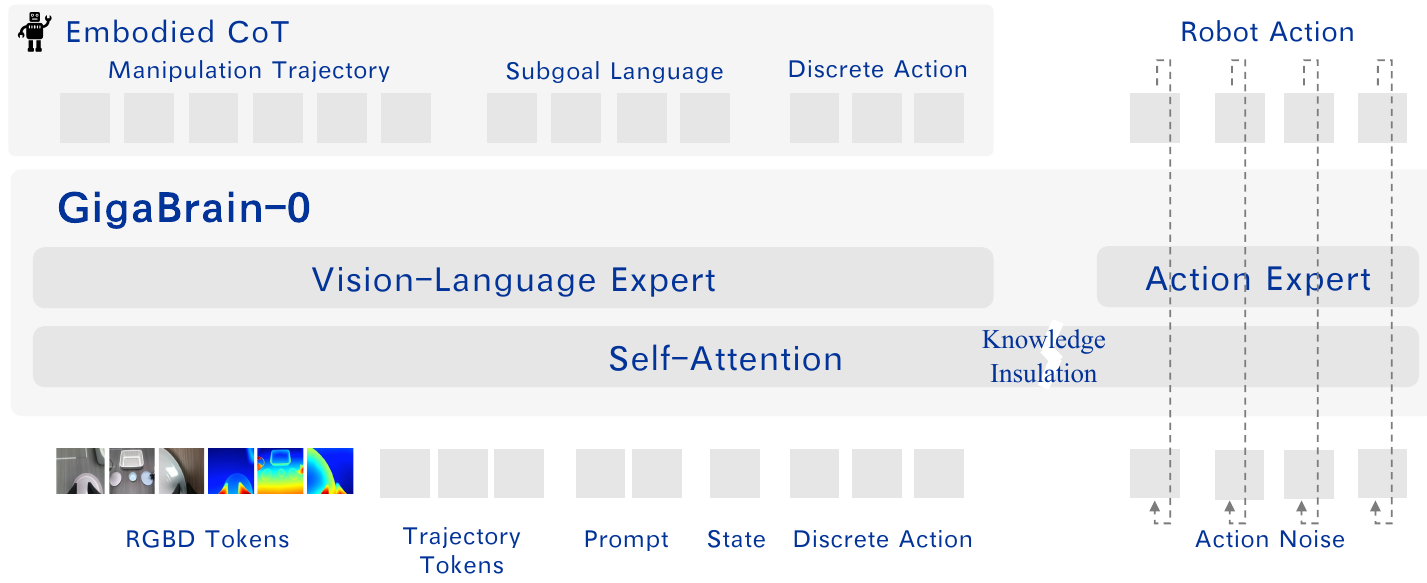}
    \captionsetup{type=figure, justification=justified, singlelinecheck=false}
    \vspace{-0.3cm}
    \caption{
    The framework of \textit{GigaBrain-0}. \textit{GigaBrain-0} takes RGB-D input to enhance spatial perception and outputs Embodied Chain-of-Thought (Embodied CoT) as an intermediate representation to strengthen embodied reasoning for manipulation. During training, \textit{GigaBrain-0} employs Knowledge Insulation~\citep{KI} to prevent interference between the optimization processes of action prediction and Embodied CoT generation.
    }
    \vspace{-0.4cm}
    \label{fig:framework}
\end{figure}

\textit{GigaBrain-0} is an end-to-end VLA model $g_\theta$ that, given visual observations and high-level language instructions, reasons over embodied scenarios to generate compliant action sequences for controlling a wheeled bi-manual robot (e.g., Agilex, G1). To enhance prompt-following fidelity and enable smoother action generation, as illustrated in Fig.~\ref{fig:framework}, \textit{GigaBrain-0} adopts a mixture-of-transformers~\citep{mot} architecture, it leverages a pretrained Vision-Language Model (VLM), PaliGemma2~\citep{paligemma2}, to encode multimodal inputs, and employs an action Diffusion Transformer (DiT)~\citep{dit} with flow matching ~\citep{lipman2022flow} to predict action chunks. This hybrid architecture enables decoupled yet synergistic processing of semantic understanding and continuous action generation. During training, we introduce Knowledge Insulation~\citep{KI} to mitigate interference between continuous action-space learning and the VLM’s semantic reasoning capabilities. Furthermore, we augment the VLM head with discrete action token prediction~\citep{pertsch2025fast}, which significantly accelerates pretraining convergence.

To enhance spatial reasoning capabilities, we incorporate RGB-D data during pretraining. Given an input tensor of shape $B \times H \times W \times 4$ (RGB + depth), we first normalize the input and extract visual features using SigLIP~\citep{siglip}. To adapt SigLIP to RGB-D inputs, we extend its first convolutional layer with zero-initialized kernels for the depth channel. This preserves pretrained RGB feature extraction while enabling depth-aware representation learning. Notably, SigLIP remains fully trainable throughout \textit{GigaBrain-0}'s training, allowing adaptive fine-tuning to embodied RGB-D perception. During training, we randomly drop the depth channel (replacing it with zero padding) to ensure compatibility with RGB-only inputs during inference.

Inspired by Chain-of-Thought (CoT) reasoning in LLMs~\citep{cot}, we introduce Embodied CoT to improve \textit{GigaBrain-0}'s reasoning in embodied environments. Unlike standard LLMs, \textit{GigaBrain-0} explicitly generates intermediate reasoning tokens, including: (1) \textit{manipulation trajectories}: 2D projections of end-effector paths onto the image plane, represented by 10 uniformly sampled keypoints; (2) \textit{subgoal language}: natural language descriptions of intermediate objectives; and (3) \textit{discrete action tokens}: discrete representations that accelerate training convergence of the subsequent DiT-based continuous action chunk prediction~\citep{pertsch2025fast}.

To balance model expressiveness with inference efficiency, we forgo autoregressive decoding for trajectory prediction. Instead, we introduce 10 learnable trajectory tokens as auxiliary inputs to the Vision-Language Model (VLM). During feature extraction, these tokens interact with the full visual context via bidirectional (non-causal) attention, enabling holistic spatial reasoning over the scene. The resulting output trajectory tokens are then passed through a lightweight GRU decoder to regress the 2D pixel-space coordinates of the end-effector’s manipulation trajectory.

In contrast, subgoal language and discrete action tokens are generated autoregressively and supervised via standard next-token prediction. All components, including trajectory regression, language-based subgoals, discrete action tokens, and continuous action chunks predicted by the Diffusion Transformer (DiT) $f_{\theta}$, are jointly optimized under a unified objective:

\begin{equation}
    \mathcal{L} = \mathbb{E}_{\mathcal{D}, \tau, \epsilon}\left[
        -\sum_{j=1}^{n-1} M_{\text{CoT},j} \log p_{\theta}\left(x_{j+1} \mid x_{1:j}\right)
        + \left\| \epsilon - a_{\text{chunk}} - f_{\theta}\left(a_{\text{chunk}}^{\tau, \epsilon}\right) \right\|^{2}
        + \lambda \left\| \text{GRU}(\hat{\mathbf{t}}_{1:10}) - \mathbf{t}_{1:10} \right\|^2
    \right],
\end{equation}
where $\mathcal{D}$ denotes the training dataset. $\tau \in [0, 1]$ is the flow-matching timestep, $\epsilon \sim \mathcal{N}(0, \mathbf{I})$ is Gaussian noise. $a_{\text{chunk}}^{\tau, \epsilon} = \tau \cdot a_{\text{chunk}} + (1 - \tau) \cdot \epsilon$ is the noised action chunk used in flow matching. $M_{\text{CoT},j} \in \{0,1\}$ is a per-token mask indicating whether position $j$ belongs to the CoT reasoning stream (subgoal language or discrete actions). $\hat{\mathbf{t}}_{1:10}$ and $\mathbf{t}_{1:10}$ denote predicted and ground-truth 2D trajectory keypoints, respectively. $\lambda=1$ is a hyperparameter balancing the trajectory regression loss. Notably, we do not manually assign loss weights to the language and action prediction terms, as Knowledge Insulation~\citep{KI} inherently prevents interference between their optimization processes, allowing each stream to learn independently.

\begin{table}[t]
\centering
\captionsetup{type=table, justification=justified, singlelinecheck=false}
\caption{Comparison of training data usage across VLA models. \textit{GigaBrain-0} leverages a diverse set of data sources to enhance generalization and reduce dependency on real-world robot data.}
\label{tab:data_support}
\begin{tabular}{l *{7}{>{\centering\arraybackslash}m{1.55cm}}}
\toprule
{Method} & 
\rot{Public Data} &
\rot{Self-Collected Data} &
\rot{Egocentric Human Data} &
\rot{Video-Gen Data} &
\rot{Sim2Real Transfer Data} &
\rot{Real2Real Transfer Data} &
\rot{View Transfer Data} \\
\midrule
$\pi_0$            & \cmark & \cmark & \xmark & \xmark & \xmark & \xmark & \xmark \\
$\pi_{0.5}$          & \cmark & \cmark & \xmark & \xmark & \xmark & \xmark & \xmark \\
G0       & \cmark & \cmark & \xmark & \xmark & \xmark & \xmark & \xmark \\
WALL-OSS       & \cmark & \cmark & \xmark & \xmark & \xmark & \xmark & \xmark \\
GR-3           & \cmark & \cmark & \cmark & \xmark & \xmark & \xmark & \xmark \\
GR00T N1.5    & \cmark & \cmark & \cmark & \cmark & \cmark & \xmark & \xmark \\
\midrule
\textbf{\textit{GigaBrain-0}}    & \cmark & \cmark & \cmark & \cmark & \cmark & \cmark & \cmark \\
\bottomrule
\end{tabular}
\end{table}

\section{GigaBrain-0 Data}
\label{sec:data}
For Vision-Language-Action (VLA) models, training data diversity is paramount~\citep{shi2025diversity}. While numerous public datasets~\citep{oxe,robomind,agibot} exist, they are often insufficient in scene variation, or task complexity to enable robust generalization. Real-world robot data collection suffers from high operational costs, low scaling efficiency, and limited environmental diversity, as most deployments repeatedly sample from the same narrow set of scenes.

Recent advances~\citep{gr00t,dreamgen} have demonstrated that world models can effectively generate diverse and photorealistic training data to augment vision-language-action (VLA) capabilities. In \textit{GigaBrain-0}, we further extend this paradigm by integrating a broad spectrum of world-model-generated data sources. As shown in Tab.~\ref{tab:data_support}, \textit{GigaBrain-0} leverages a more diverse set of data sources compared to existing VLA counterparts~\citep{pi0,pi05,walloss,galaxea,gr3,gr00t}. This expanded data diversity significantly reduces reliance on real-world robot-collected data while enhancing model generalization. We detail each data source in the following.

\begin{figure}[htbp]
    \centering
    \includegraphics[width=1\linewidth]{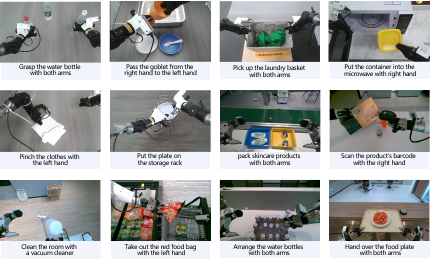}
    \captionsetup{type=figure, justification=justified, singlelinecheck=false}
    \caption{\textit{GigaBrain-0}'s self-collected real-world robot data is gathered from PiPER arms and the AgiBot G1 platform, spanning diverse environments including homes, supermarkets, factories, and office settings.}
    \label{fig:real_data}
\end{figure}

\subsection{Real-World Data}
\label{subsec:real_data}

\textbf{Data Source}. Our real-world data corpus integrates both publicly available datasets and proprietary data collected from our in-house robotic platforms. Publicly sourced datasets include AgiBotWorld~\citep{agibot}, RoboMind~\citep{robomind}, and Open X-Embodiment~\citep{oxe}, which collectively provide foundational coverage of manipulation and locomotion tasks. 
In addition, we collect 1182 hours proprietary data using Agilex Cobot Magic platform (199 hours) and the AgiBot G1 platform (983 hours) across a total area of 3100$\text{m}^2$, spanning five broad environment categories: industrial, commercial, office, residential, and laboratory settings. These are further subdivided into 14 distinct real-world scenarios, including supermarkets, hotel lobbies, coffee shops, bubble tea stores, convenience stores, restaurants, warehouse material handling, industrial assembly lines, pantries, private residences, apartment interiors, meeting rooms, office workstations, and laboratories.
The collected tasks range from basic pick-and-place operations to long-horizon sequential activities, mobile manipulation in dynamically changing layouts, and interactions with deformable objects, as illustrated in Fig.~\ref{fig:real_data}.

\textbf{Data Annotation \& Processing}. 
For data annotation, if the captured RGB frames lack depth information, we employ MoGe~\citep{moge} to generate metrically scaled depth maps. Regarding subgoal language annotation, we observe that VLMs struggle to accurately segment long-horizon tasks into meaningful subgoals, while manual segmentation proves prohibitively time-consuming. To address this, we adopt an approach inspired by~\citep{rlbench}, leveraging gripper state transitions (e.g., open/close, grasp/release) to automatically segment trajectories into atomic subtasks.
For each segmented subtask, we utilize Qwen-VL-2.5~\citep{qwen25vl} to generate subgoal language annotations. To mitigate hallucination and ensure consistency, we constrain the annotation process using a structured template and a predefined vocabulary of standardized action phrases (e.g., \texttt{pick [object]}, \texttt{place [object] into [container]}, \texttt{open [device]}), selected from a curated description library.
For 2D manipulation trajectory annotation, we project the 3D end-effector coordinates into the head-mounted camera’s image plane, yielding pixel-space motion traces aligned with visual observations.
Notably, we annotate only a subset of the collected data and train our models using a mixture of fully annotated, partially annotated, and raw unannotated trajectories to maximize data utility while managing annotation costs.
To further improve pretraining efficiency and reduce redundancy, we perform deduplication across the entire corpus. For each unique task, we retain at most 50 diverse demonstration trajectories, preserving behavioral variety while eliminating near-identical repetitions. This strategy enhances sample efficiency and promotes more robust and generalizable model learning.

\subsection{World-Model-Generated Data}
\label{subsec:synthetic_data}

To overcome the limitations of physical data collection, we employ GigaWorld, our world-model framework, to generate diverse, physically plausible training sequences. GigaWorld synthesizes data through multiple complementary pipelines:

\begin{figure}[htbp]
\centering
\captionsetup{type=figure, justification=justified, singlelinecheck=false}
\includegraphics[width=1\linewidth]{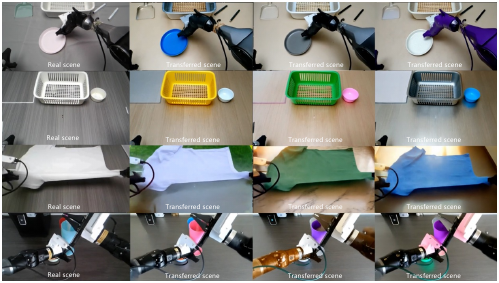}
\caption{GigaWorld enables Real2Real apperance transfer by taking real-world captured data and generating generalized variations in texture, color, lighting, and material properties.}
\label{fig:real2real}
\end{figure}

\textbf{Real2Real Transfer}. Real-world robot-collected videos are inherently constrained by the physical environments in which they are captured, including static backgrounds, fixed lighting conditions, and limited variation in object materials, textures, and colors. To overcome these limitations and significantly enhance visual and contextual diversity, we leverage GigaWorld to perform Real2Real Transfer: re-rendering real trajectories in synthetically altered but physically plausible visual contexts. Specifically, we train diffusion-based video generation model~\citep{emma,robotransfer} conditioned on geometric and structural priors extracted from real footage. We employ VideoDepthAnything~\citep{videodepth} to estimate per-frame depth maps and extract Canny edge maps to preserve object boundaries and scene structure. These signals serve as spatial control conditions via a ControlNet~\citep{controlnet} branch, enabling precise manipulation of appearance while preserving motion and layout consistency. During inference, for each real-world video clip, we generate approximately 10 visually distinct variants by textually prompting changes to foreground/background materials, surface textures, illumination conditions, and color palettes, all while maintaining the original action semantics and spatial dynamics, as shown in Fig.~\ref{fig:real2real}. This approach efficiently multiplies the effective diversity of real data without requiring additional physical collection.

\begin{figure}[htbp]
\centering
\captionsetup{type=figure, justification=justified, singlelinecheck=false}
\includegraphics[width=1\linewidth]{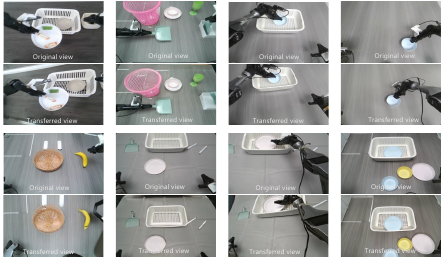}
\caption{GigaWorld supports view transfer by re-rendering real-world captured data from diverse viewpoints, thereby enriching the dataset with varied perspective changes.}
\label{fig:view_transfer}
\end{figure}

\textbf{View Transfer}.
Beyond texture and illumination variation, generalization across observation viewpoints is equally critical for robust embodied perception~\citep{xing2025shortcut}. To this end, we exploit GigaWorld’s viewpoint generalization capability to augment real robot-collected data with novel camera perspectives while preserving 3D scene consistency.
Specifically, to ensure geometric coherence under viewpoint changes, we project the original RGB frames into novel views using collected depth maps. If depth is unavailable in the source data, we annotate metric-scale depth using MoGe~\citep{moge}. The reprojected views inevitably contain occluded or incomplete regions, we inpaint these using a DiT-based video completion model~\citep{egodemogen} conditioned on the reprojected views.
Notably, when the camera viewpoint shifts, the robot’s end-effector must remain functionally consistent with the task, even though its joint configuration changes. We compute the new joint angles via IK based on the updated ego-pose and end-effector pose. The resulting articulated robot geometry is then rendered using its URDF model in a physics-aware simulation engine, and provided as a structural condition~\citep{egodemogen} to the generative model.
To mitigate potential discrepancies between simulated and real robot dynamics, we optionally employ a differentiable physics engine~\citep{embodiedreamer} to fine-tune motion plausibility and close the sim-to-real physics gap.
As shown in Fig.~\ref{fig:view_transfer}, from a single real-world trajectory, our pipeline generates multiple viewpoint-consistent renderings of the scene, complete with dynamically adjusted robot poses that preserve task semantics and physical feasibility.

\begin{figure}[htbp]
\centering
\captionsetup{type=figure, justification=justified, singlelinecheck=false}
\includegraphics[width=1\linewidth]{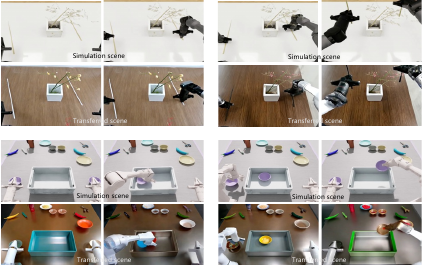}
\caption{GigaWorld enables Sim2Real transfer by generalizing simulation-collected data in terms of texture, color, lighting, and material properties to better bridge the domain gap and enhance realism.}
\label{fig:sim2real}
\end{figure}

\textbf{Sim2Real Transfer}.
While the methods described above augment real-world data, we further scale our training corpus by leveraging simulation assets to synthesize diverse embodied interaction sequences. Specifically, we compose manipulation scenes in Isaac Sim~\citep{issacsim} using either procedurally generated assets from EmbodiedGen~\citep{embodiedgen} or curated objects from open-source repositories such as ArtVIP~\citep{artvip}. Robot morphology is defined via URDF files, and end-effector trajectories are computed using IK to ensure physically plausible motion.
To bridge the sim-to-real domain gap, particularly in visual appearance, we apply GigaWorld’s Sim2Real Transfer pipeline. Similar in architecture to Real2Real Transfer, this method conditions a diffusion-based video generator~\citep{robotransfer,emma} on depth maps exported from the simulation environment. Using text prompts, we dynamically alter surface textures, material reflectance, lighting conditions, and environmental clutter to produce photorealistic renderings that retain the original scene geometry and action semantics, as illustrated in Fig.~\ref{fig:sim2real}.
Crucially, unlike real-world data, simulation affords us full control over scene parameters: we can systematically vary object initial positions, camera viewpoints, background layouts, and even physics properties (e.g., friction, mass) to maximize combinatorial diversity.

\begin{figure}[t]
\centering
\captionsetup{type=figure, justification=justified, singlelinecheck=false}
\includegraphics[width=1\linewidth]{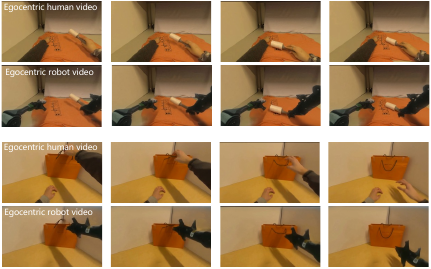}
\caption{GigaWorld supports egocentric human video transfer by transforming first-person human hand actions into robotic manipulation scenarios, effectively mapping human demonstrations to robot-executable tasks.}
\label{fig:human_transfer}
\end{figure}

\textbf{Human Video Transfer}. 
Human demonstration videos have emerged as a promising source for training embodied agents~\citep{gr3,mimicplay,egovla,egomimic,bu2025univla}, offering diversity in tasks, environments, and interaction styles, far exceeding the scale and variety achievable through robotic data collection alone. However, raw human videos exhibit significant gaps when directly applied to robot learning, as egocentric footage often suffers from motion blur, unstable viewpoints, and a visual and kinematic mismatch between human hands and robotic end-effectors.
To bridge this gap, we leverage GigaWorld’s video inpainting capabilities to transform large-scale first-person human videos into stable, robot-centric demonstrations. Specifically, we convert videos from the EgoDex dataset~\citep{egodex} into stabilized, robot-executable sequences with articulated mechanical arms replacing human hands. Specifically, we use SAM2~\citep{sam2} to segment and mask out human hands in each frame. The annotated 3D hand wrist positions (provided in EgoDex) are treated as target end-effector poses for a simulated robot arm. We solve for corresponding joint angles via IK, then render the robot’s URDF model using a physics-aware simulation engine. This rendered arm geometry serves as a structural condition for our diffusion-based generator~\citep{mimicdreamer}, ensuring kinematically plausible and visually consistent robot appearances.
As shown in Fig.~\ref{fig:human_transfer}, the output is a stabilized, robot-embodied version of the original human demonstration, preserving task intent and spatial relationships while eliminating visual and kinematic domain gaps.

\begin{figure}[t]
\centering
\captionsetup{type=figure, justification=justified, singlelinecheck=false}
\includegraphics[width=1\linewidth]{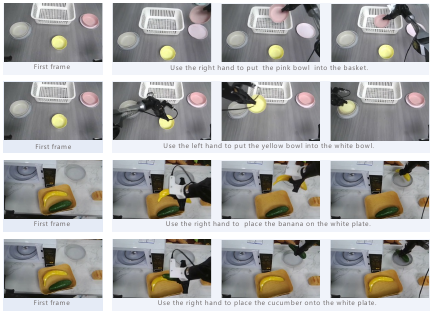}
\caption{GigaWorld can generate diverse future trajectories from the same initial frame under different text prompts, thereby augmenting the dataset with novel manipulation sequences.}
\label{fig:videogen}
\end{figure}

\textbf{Video Generation with Inverse Dynamics Modeling}.
Given a single input image, GigaWorld can generate diverse embodied robotic operation videos conditioned on different textual prompts, as illustrated in Fig.~\ref{fig:videogen}. Furthermore, we leverage Inverse Dynamics Models (IDM)~\citep{dreamgen} to infer corresponding robot action sequences from these generated videos, which are then used as synthetic training data for embodied manipulation tasks.

\begin{figure}[t]
\centering
\captionsetup{type=figure, justification=justified, singlelinecheck=false}
\includegraphics[width=1\linewidth]{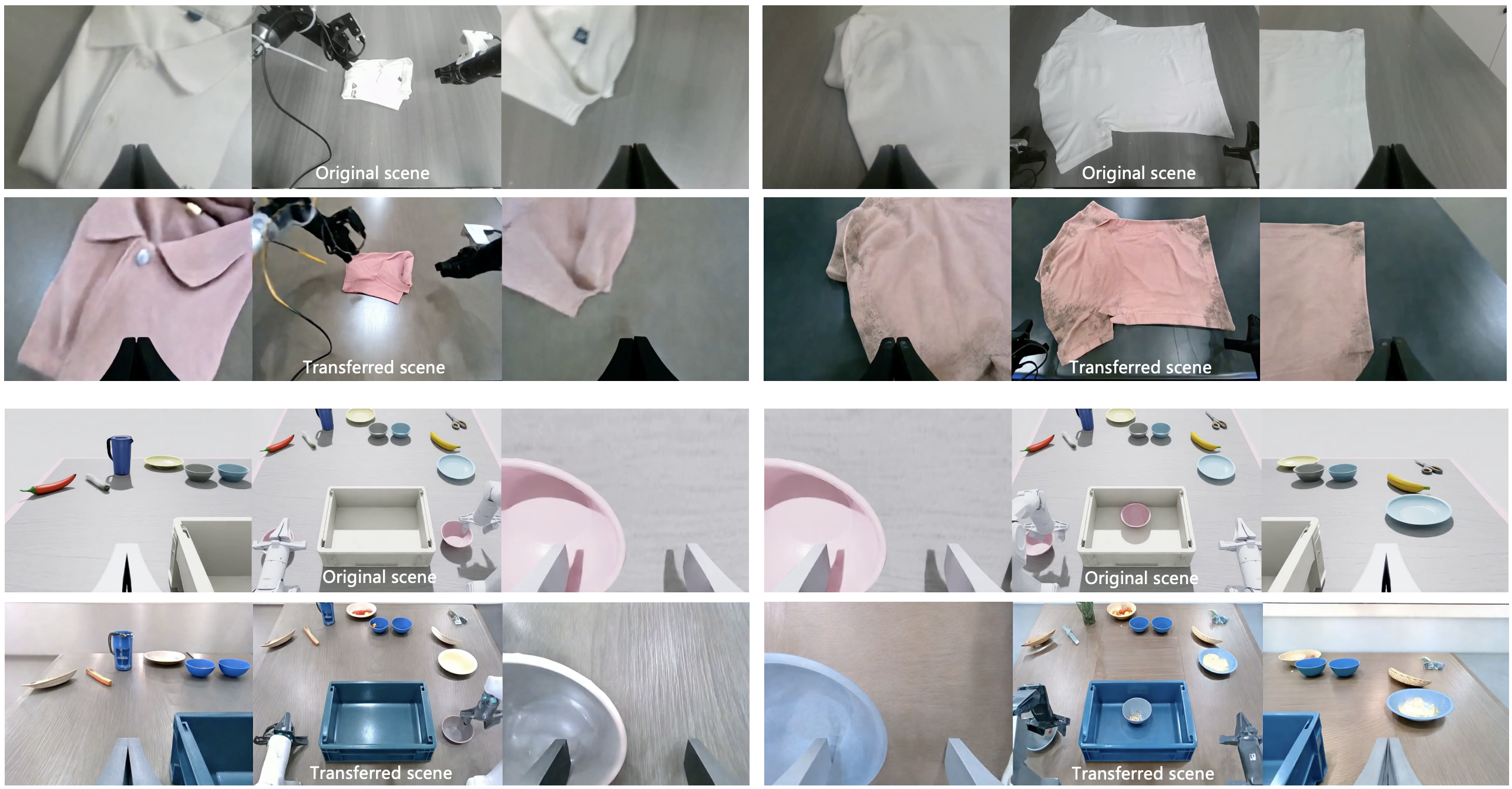}
\caption{GigaWorld can generate multi-view consistent videos, thereby enabling 3D-aware training and improving spatial reasoning in downstream tasks.}
\label{fig:mv}
\end{figure}

\textbf{Multiview Video Generation}.
In embodied manipulation scenarios, multiple cameras (e.g., head-mounted and wrist-mounted cameras) are often deployed to capture the operating environment from different viewpoints, creating a demand for generating temporally and spatially consistent multiview videos. To address this, GigaWorld adopts the multiview modeling approach from~\citep{drivedreamer2,robotransfer,emma}, which concatenates noise maps from multiple views as input to the diffusion model. This design preserves the original diffusion architecture without modification and enables consistent multiview video generation with only minimal fine-tuning data. As shown in Fig.~\ref{fig:mv}, GigaWorld is capable of generating diverse yet geometrically consistent multiview videos, demonstrating strong cross-view coherence and scene fidelity.

\textbf{Generation Efficiency}.
Video diffusion models~\citep{cosmos,hunyuanvideo,wan} suffer from low generation efficiency, often requiring tens of minutes to synthesize hundreds of frames resolution. To accelerate inference, GigaWorld employs NATTEN~\citep{natten} as a computationally efficient alternative to standard self-attention. Furthermore, GigaWorld leverages step distillation~\citep{dmd2} to reduce the denoising process from dozens of steps to a single-step generation. Combined with FP8-precision inference, these optimizations collectively yield over a 50$\times$ speedup in video generation compared to baseline diffusion models.

\textbf{Generated Data Quality Inspection}.
Generated videos inevitably contain hallucinations or artifacts that may degrade downstream training performance. To mitigate this, GigaWorld introduces a comprehensive quality assessment pipeline that evaluates generated videos across multiple dimensions: geometric consistency~\citep{robotransfer}, multiview consistency~\citep{robotransfer}, text-description alignment~\citep{cosmosreason}, and physical plausibility~\citep{cosmosreason}. Each video is assigned a composite quality score, which determines whether it is suitable for pre-training, fine-tuning, or should be discarded. All model and training details mentioned above will be fully elaborated in the upcoming GigaWorld technical report.

\section{Experiment}

To evaluate the real-world manipulation performance, we conducted robot experiments on two robotic platforms: the dual-arm PiPER robot platform and the AgiBot G1 platform. These experiments covered a diverse range of tasks, including dexterous manipulation (e.g., \texttt{laundry folding} and \texttt{paper towel preparation}), long-horizon tasks (e.g., \texttt{table bussing} and \texttt{juice preparation}), and mobile manipulation (e.g., \texttt{boxes moving} and \texttt{laundry basket moving}). Furthermore, we validated \textit{GigaBrain-0}'s generalization capabilities, specifically in appearance, placement, and viewpoint generalization. We also performed on-device experiments to verify the real-time inference performance of the lightweight variant, \textit{GigaBrain-0}-Small. In the following sections, we will elaborate on the experimental setup and present detailed performance comparisons.

\begin{figure}[htbp]
\centering
\captionsetup{type=figure, justification=justified, singlelinecheck=false}
\includegraphics[width=1\linewidth]{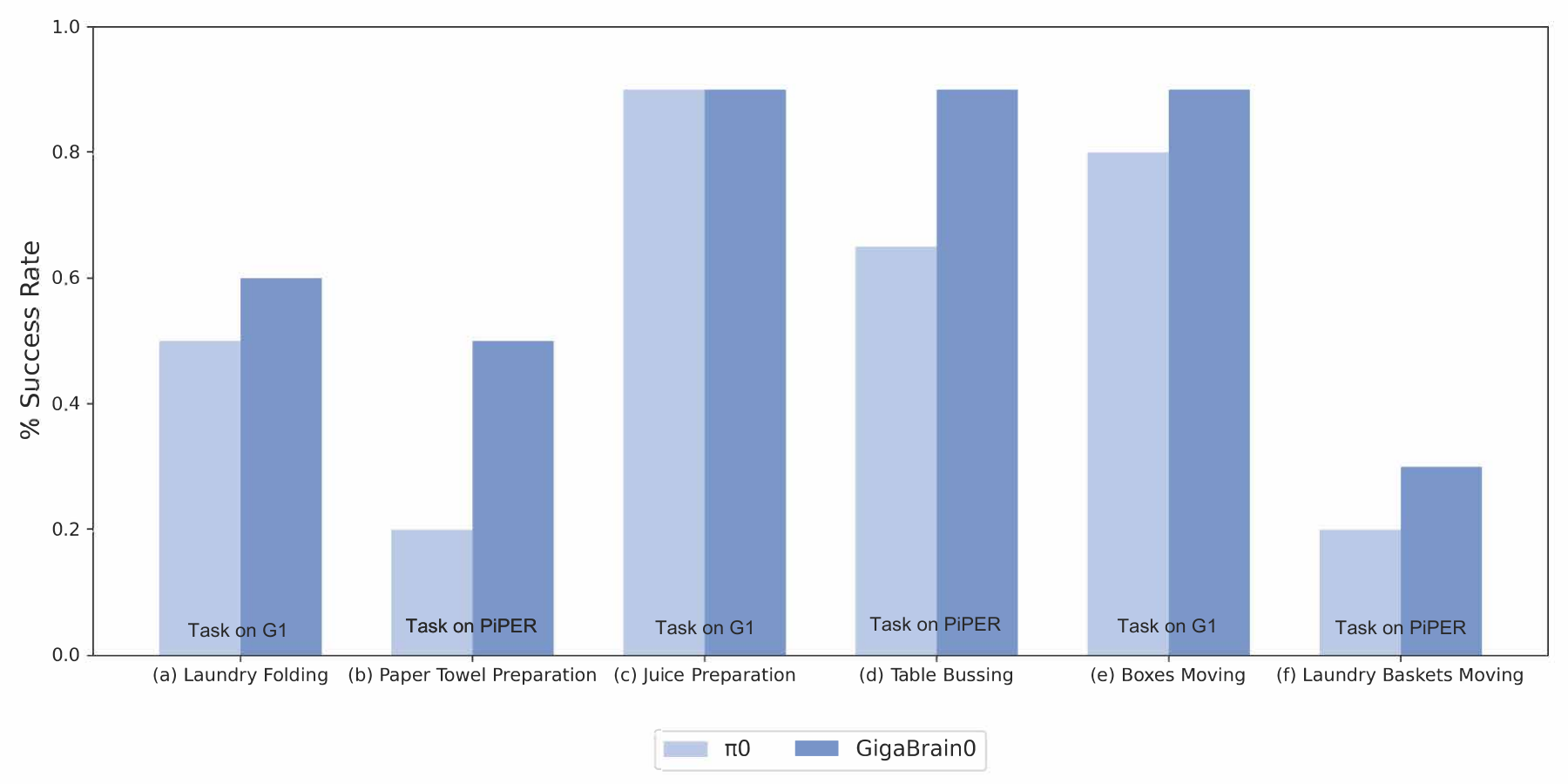}
\caption{Performance comparison between \textit{GigaBrain-0} and $\pi_0$ across six tasks on G1 and PiPER robot platforms, where (a) \texttt{Laundry Folding} and (b) \texttt{Paper Towel Preparation} are dexterous manipulation tasks; (c) \texttt{Juice Preparation} and (d) \texttt{Table Bussing} are long-horizon tasks; (e) \texttt{Boxes Moving} and (f) \texttt{Laundry Baskets Moving} are mobile manipulation tasks.}
\label{fig:main_exp}
\end{figure}

\subsection{Dexterous Manipulation Experiment}

\textbf{Experiment Setup.}
We evaluate \textit{GigaBrain-0} on two dexterous manipulation tasks, \texttt{laundry folding} and \texttt{paper towel preparation}. For \texttt{laundry folding}, we deploy the model on the G1 robot platform and fine-tune it using 300 human-collected demonstration trajectories, the training is conducted with a batch size of 128 for 40K steps. For \texttt{paper towel preparation}, we use 100 demonstrations from PiPER arms, and the training is conducted with a batch size of 128 for 20K steps.
We compare \textit{GigaBrain-0} against $\pi_0$~\citep{pi0}, which is implemented using the official open-source code, and fine-tuned with the same training config for fair comparison.


\begin{figure}[htbp]
\centering
\captionsetup{type=figure, justification=justified, singlelinecheck=false}
\includegraphics[width=1\linewidth]{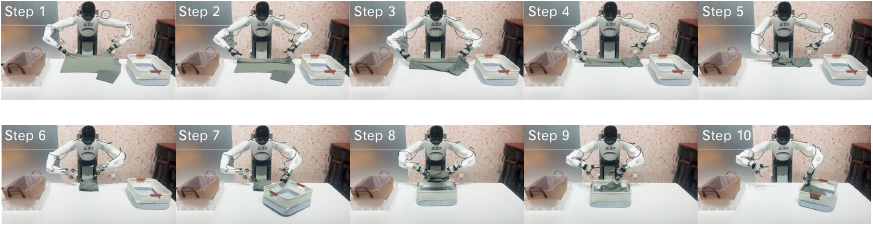}
\caption{Deployment of \textit{GigaBrain-0} on the G1 humanoid robot for real-world \texttt{laundry folding}.}
\label{fig:laundry_demo}
\end{figure}

\begin{figure}[htbp]
\centering
\captionsetup{type=figure, justification=justified, singlelinecheck=false}
\includegraphics[width=1\linewidth]{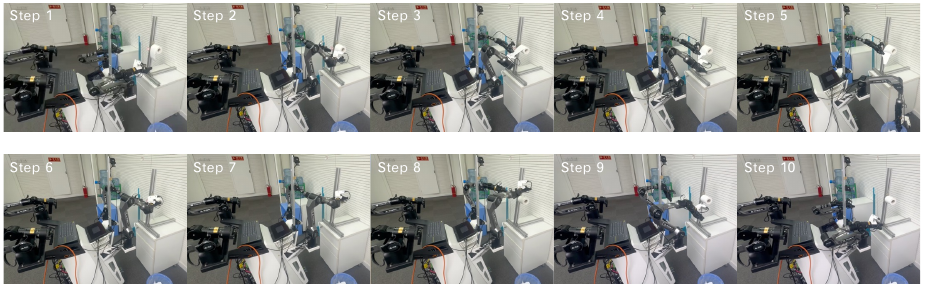}
\caption{Deployment of \textit{GigaBrain-0} on the PiPER arms for real-world \texttt{paper towel preparation}.}
\label{fig:papertowel_demo}
\end{figure}

\textbf{Experiment Results.}
As shown in Fig.~\ref{fig:main_exp}(a--b), \textit{GigaBrain-0} consistently achieves the highest task success rate across both tasks, surpassing $\pi_{0}$ by 30\% and 10\% in each case. Its integration of depth sensing enhances spatial awareness, enabling precise coordination in contact-rich scenarios. Qualitatively, as illustrated in Fig.~\ref{fig:laundry_demo} and Fig.~\ref{fig:papertowel_demo}, \textit{GigaBrain-0} robustly executes complex, dexterous workflows: for laundry, it performs synchronized dual-gripper grasping, accurate folding, and final alignment; for paper towels, it tears off excess material, rolls the towel with controlled tension, and precisely applies adhesive labels. These results highlight \textit{GigaBrain-0}'s superior dexterous manipulation capability and generalization in real-world, unstructured settings.

\subsection{Long-horizon Experiment}

\textbf{Experiment Setup.}
We evaluate \textit{GigaBrain-0} on two long-horizon manipulation tasks: \texttt{table bussing} and \texttt{juice preparation}. For \texttt{table bussing}, we deploy the model on the dual-arm PiPER robot platform and fine-tune it using 100 human-collected demonstration trajectories, and the training is conducted with a batch size of 128 for 20K steps. For \texttt{juice preparation}, we utilize the AgiBot G1 robot with 489 demonstrations, and the training is conducted with a batch size of 128 for 35K steps. We compare \textit{GigaBrain-0} against $\pi_{0}$~\citep{pi0}, which is implemented using the official open-source code, and fine-tuned with the same training config to ensure a fair comparison.

\begin{figure}[htbp]
\centering
\captionsetup{type=figure, justification=justified, singlelinecheck=false}
\includegraphics[width=1\linewidth]{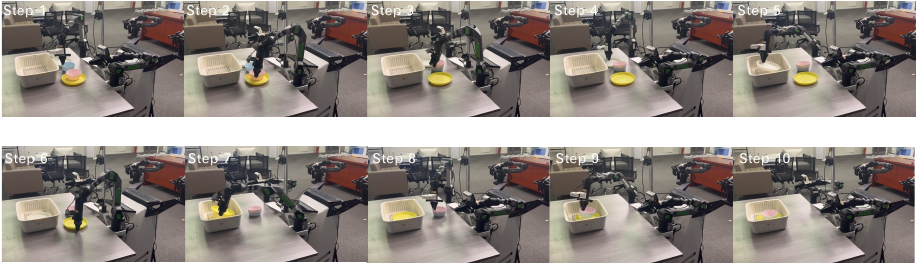}
\caption{Deployment of \textit{GigaBrain-0} on PiPER arms for real-world \texttt{table bussing}.}
\label{fig:table_demo}
\end{figure}

\begin{figure}[htbp]
\centering
\captionsetup{type=figure, justification=justified, singlelinecheck=false}
\includegraphics[width=1\linewidth]{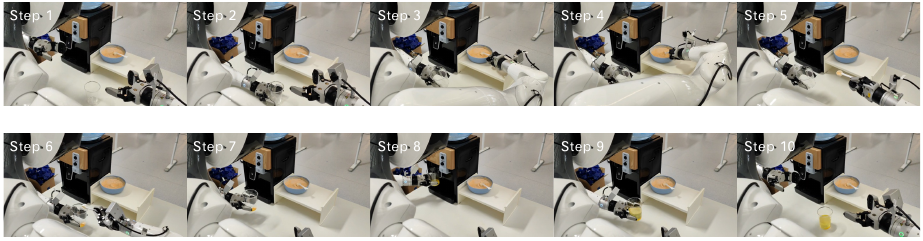}
\caption{Deployment of \textit{GigaBrain-0} on G1 humanoid robot for real-world \texttt{juice preparation}.}
\label{fig:juice_demo}
\end{figure}

\textbf{Experiment Results.}
As shown in Fig.~\ref{fig:main_exp}(c--d), \textit{GigaBrain-0} consistently achieves the highest task success rate across both long-horizon tasks. Its integration of embodied Chain-of-Thought (CoT) reasoning enables fine-grained, temporally ordered planning—critical for complex multi-step execution. Qualitative results in Fig.~\ref{fig:table_demo} and Fig.~\ref{fig:juice_demo} further demonstrate \textit{GigaBrain-0}'s robustness: in \texttt{table bussing}, it first detaches bowls from plates, then sequentially places plates followed by bowls into the collection bin. In \texttt{juice preparation}, it scoops powdered mix with a spoon, dispenses water from the dispenser, and finally stirs the mixture to completion. These results highlight \textit{GigaBrain-0}'s superior capacity for structured reasoning and reliable execution in long-horizon, real-world manipulation scenarios.

\subsection{Mobile Manipulation Experiment}

\textbf{Experiment Setup.}
We evaluate \textit{GigaBrain-0} on two mobile manipulation tasks: \texttt{boxes moving} and \texttt{laundry baskets moving}. For \texttt{boxes moving}, we deploy the model on the AgiBot G1 robot platform and fine-tune it using 300 human-collected demonstration trajectories, and the training is conducted with a batch size of 128 for 30K steps. For \texttt{laundry baskets moving}, we utilize the dual-arm PiPER robot platform with 378 demonstrations, and the training is conducted with a batch size of 192 for 30K steps. We compare \textit{GigaBrain-0} against $\pi_0$~\citep{pi0}, which is implemented using the official open-source code, and fine-tuned with the same training config to ensure a fair comparison.

\begin{figure}[htbp]
\centering
\captionsetup{type=figure, justification=justified, singlelinecheck=false}
\includegraphics[width=1\linewidth]{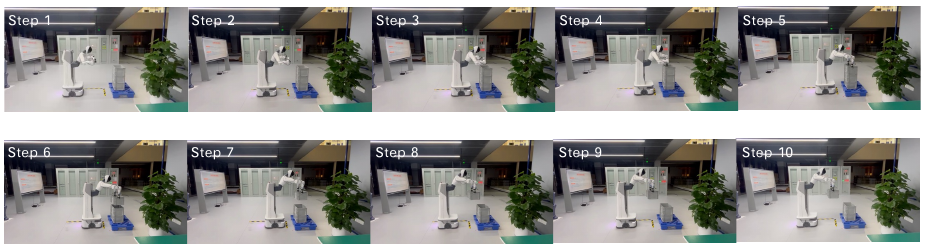}
\caption{Deployment of \textit{GigaBrain-0} on the G1 humanoid robot for real-world \texttt{paper towel preparation}.}
\label{fig:boxes_demo}
\end{figure}

\begin{figure}[htbp]
\centering
\captionsetup{type=figure, justification=justified, singlelinecheck=false}
\includegraphics[width=1\linewidth]{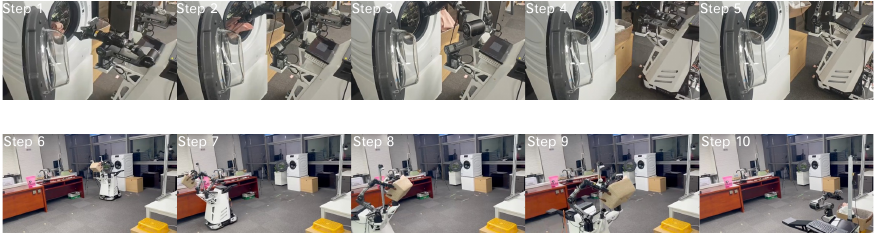}
\caption{Deployment of \textit{GigaBrain-0} on the PiPER arms for real-world \texttt{laundry baskets moving}.}
\label{fig:baskets_demo}
\end{figure}

\textbf{Experiment Results.}
As shown in Fig.~\ref{fig:main_exp}(e--f), \textit{GigaBrain-0} achieves the highest success rate in both mobile manipulation tasks, surpassing $\pi_{0}$  by 10\% in each task. Its unified architecture, combining global navigation priors with local manipulation policies, enables seamless transitions between mobility and interaction. In \texttt{boxes moving}, \textit{GigaBrain-0} navigates around obstacles, identifies target boxes using semantic segmentation, grasps them with adaptive force control, and delivers them to designated zones. In \texttt{laundry baskets moving}, it locates partially occluded baskets, adjusts its base position for optimal reach, lifts baskets using compliant gripper control, and transports them across uneven terrain without spilling contents. Qualitative results in Fig.~\ref{fig:boxes_demo} and Fig.~\ref{fig:baskets_demo} further illustrate \textit{GigaBrain-0}'s robustness to environmental variability and its ability to recover from minor localization or grasping errors through real-time replanning.

\begin{figure}[t]
\centering
\captionsetup{type=figure, justification=justified, singlelinecheck=false}
\includegraphics[width=1\linewidth]{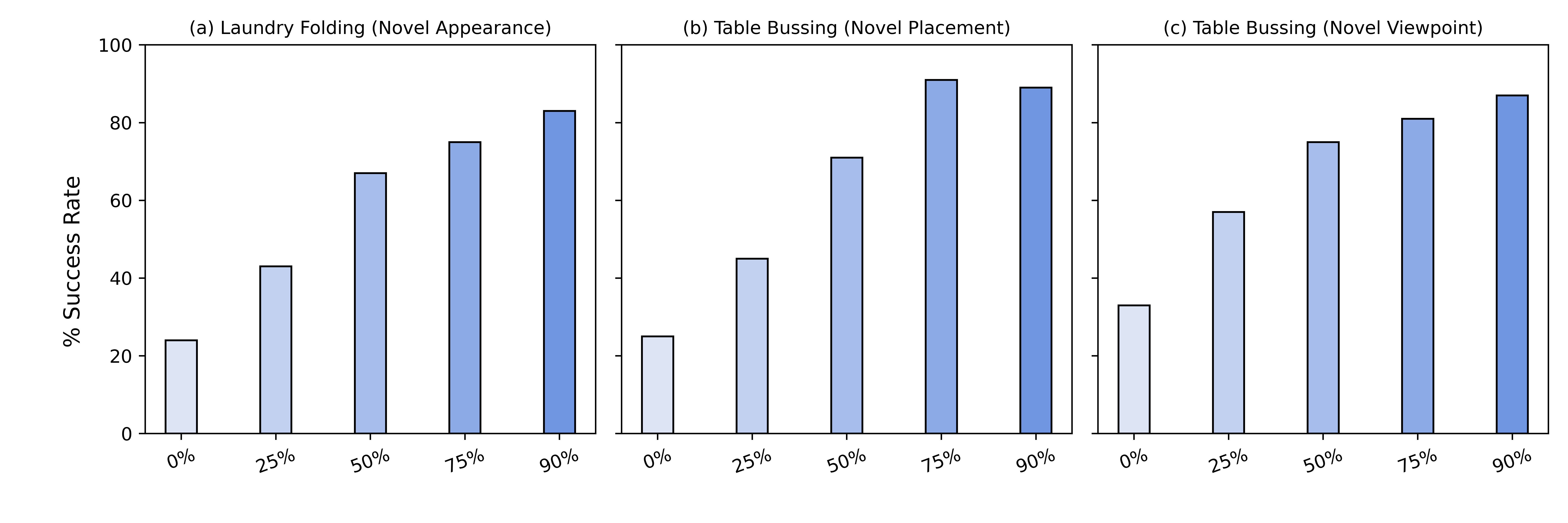}
\caption{Generalization performance of \textit{GigaBrain-0} under appearance, placement, and viewpoint shifts. The horizontal axis denotes the sampling probability $\alpha$  of world model-generated data used during training. }
\label{fig:generalization_exp}
\end{figure}

\subsection{Manipulation Generalization Experiment}
Diverse training data is crucial for the generalization of VLA models~\citep{shi2025diversity}. \textit{GigaBrain-0} leverages GigaWorld to generate diverse training data, which significantly enhances its robustness and generalization under variations in appearance, object placement, and camera viewpoint. Below, we present our experimental setup and analysis in detail.

\begin{figure}[htbp]
\centering
\captionsetup{type=figure, justification=justified, singlelinecheck=false}
\includegraphics[width=1\linewidth]{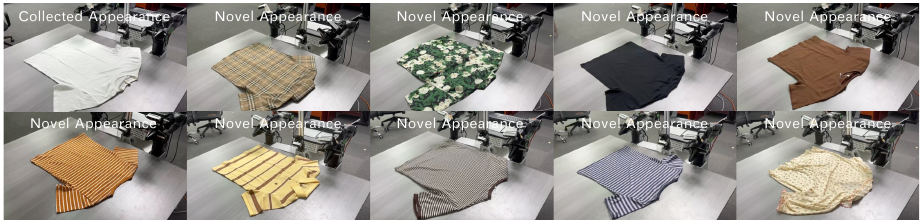}
\caption{Illustration of the appearance generalization experiment on \texttt{laundry folding}: one white garment and nine items with varied colors and textures.}
\label{fig:appearance}
\end{figure}

\textbf{Appearance Generalization.} We conduct appearance generalization experiments on the \texttt{laundry folding} task. Specifically, we perform post-training on \textit{GigaBrain-0} using a small set of real-world folding demonstrations, limited to only 50 trajectories of white garments collected on the physical robot. To enhance visual diversity, we combine this real data with trajectories generated by the GigaWorld Real2Real transfer model, which effectively transfers appearance attributes while preserving geometric and dynamic consistency (see Fig.~\ref{fig:real2real} for illustration).
We then evaluate how the mixing ratio $\alpha$ (i.e., the probability of sampling generated data in each training batch) affects real-world success rates. The model is fine-tuned for 20k steps with a batch size of 128. For evaluation, we test on 10 distinct garments with diverse appearances, including the original white garments and 9 additional items with varied colors and textures (examples shown in Fig.~\ref{fig:appearance}). Each garment is folded five times, and success is measured per trial. 
As shown in Fig.~\ref{fig:generalization_exp}(a), for a complex deformable manipulation task like garment folding, training exclusively on white garments leads to poor generalization to real-world garments with diverse textures and colors. However, incorporating appearance-transferred data significantly improves texture generalization. Notably, as the sampling probability $\alpha$ of generated data increases from 0 to 50\%, the success rate rises substantially, reaching nearly 70\%. Further increasing $\alpha$ to 75\% and 90\% yields additional gains, pushing the success rate above 80\%.

\begin{figure}[htbp]
\centering
\captionsetup{type=figure, justification=justified, singlelinecheck=false}
\includegraphics[width=1\linewidth]{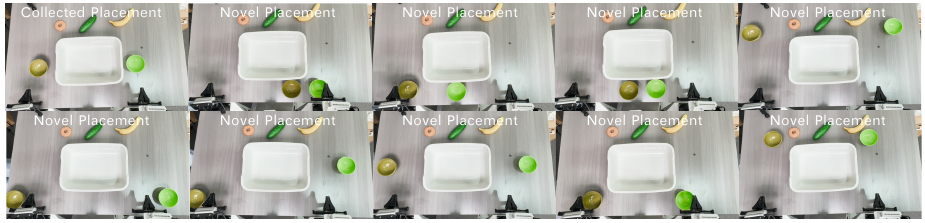}
\caption{Illustration of the object placement generalization experiment on \texttt{table bussing}: one collected placement and nine novel object placements.}
\label{fig:placement}
\end{figure}

\textbf{Placement Generalization.} We investigate generalization to novel object placements in the table bussing task. Our approach begins with post-training \textit{GigaBrain-0} on a small set of real-world demonstrations, only 50 trajectories collected under a single, fixed bowl configuration. To broaden the distribution of object positions during training, we supplement this limited real data with trajectories produced by the GigaWorld Sim2Real transfer model. This model not only introduces a wide variety of object placements but also enhances visual realism by adapting simulated appearances to better match real-world conditions (see Fig.~\ref{fig:sim2real}).
We study the impact of the mixing ratio $\alpha$ on downstream performance. The policy is fine-tuned for 20k steps with a batch size of 128. During evaluation, we assess performance across 10 distinct placement layouts: the original fixed setup with 9 new configurations with diverse object placements (examples in Fig.~\ref{fig:placement}). Each layout is tested five times, and success is recorded per trial.
Results in Fig.~\ref{fig:generalization_exp}(b) reveal that, for this long-horizon bussing task, models trained solely on fixed-placement data struggle to handle real-world variations in object layout. In contrast, blending in Sim2Real transferred data dramatically boosts generalization to unseen placements. As $\alpha$ increases from 0 to 75\%, the success rate climbs sharply, surpassing 90\%. This gain arises because the simulation provides abundant novel placement scenarios that teach the policy manipulation behaviors, while the Sim2Real transfer simultaneously mitigates the visual domain gap between simulation and real images.

\begin{figure}[htbp]
\centering
\captionsetup{type=figure, justification=justified, singlelinecheck=false}
\includegraphics[width=1\linewidth]{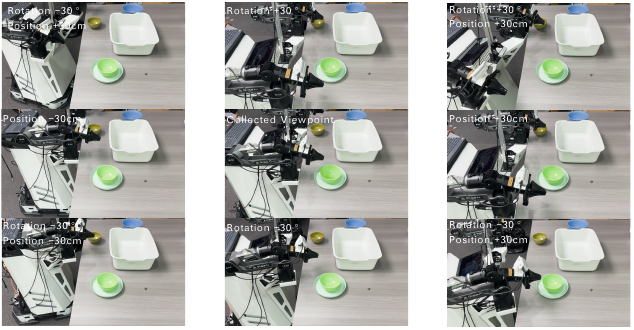}
\caption{Illustration of the camera viewpoint generalization experiment on \texttt{table bussing}: one collected viewpoint and eight novel camera viewpoints.}
\label{fig:view}
\end{figure}

\textbf{Viewpoint Generalization.} We investigate generalization to novel camera viewpoints in the table bussing task. Begining by post-training \textit{GigaBrain-0} on a small set of 50 real-world demonstrations, all captured from a single, fixed camera viewpoint. To enrich viewpoint diversity during training, we augment this data with trajectories generated by the GigaWorld view transfer model, which effectively re-renders the originally collected demonstrations from a broad range of virtual camera angles while preserving task-relevant robot actions (see Fig.~\ref{fig:view_transfer}).
We analyze how the mixing ratio $\alpha$ influences real-world performance. The policy is fine-tuned for 20k steps with a batch size of 128. For evaluation, we test on 9 distinct camera viewpoints: the original fixed view plus 8 additional, previously unseen viewpoints with varied positions and rotations (examples shown in Fig.~\ref{fig:view}). Each viewpoint is evaluated over five independent trials, with success judged per trial.
As shown in Fig.~\ref{fig:generalization_exp}(c), when trained only on data from a single viewpoint, the policy exhibits significant performance degradation under novel viewing conditions. In contrast, incorporating view-transferred data substantially improves robustness to viewpoint shifts. As $\alpha$ increases from 0 to 90\%, the success rate rises sharply, exceeding 80\%. This improvement is attributed to two factors: (1) exposure to multi-view observations enables the policy to learn viewpoint-invariant visual representations, and (2) the GigaWorld view transfer model provides photorealistic, geometrically consistent renderings that closely approximate real-world visual input from arbitrary angles.

\subsection{On-device Experiment}
Recent VLA models~\citep{gr00t,gr3,walloss,pi0,pi05,galaxea}, while powerful, are often hindered by their large parameter counts and high computational complexity, making them impractical for deployment on resource-constrained edge devices. However, efficient on-device execution is crucial for real-time robotic applications that require low latency, privacy, and autonomy. To address this challenge, we present \textit{GigaBrain-0-Small}, an optimized lightweight variant specifically designed for efficient inference on edge platforms such as the NVIDIA Jetson AGX Orin.

Compared to \textit{GigaBrain-0}, \textit{GigaBrain-0-Small} adopts the compact vision-language model SmolVLM2~\citep{smolvlm} and reduces the action expert parameters to approximately 100M. Beyond architectural simplification, we implement a series of system-level optimizations: (1) we eliminate redundant CPU–GPU memory transfers and unnecessary data type (dtype) conversions; (2) we enable automatic mixed-precision inference via \texttt{torch.autocast}; (3) we optimize the Rotary Position Embedding (RoPE)~\citep{su2024roformer} computation by precomputing and caching the sine and cosine lookup tables; and (4) we apply \texttt{torch.compile} to key components, including the denoising step and VLM forward pass, to convert dynamic PyTorch code into optimized static graphs.

These optimizations collectively enable \textit{GigaBrain-0-Small} to achieve significantly lower latency and memory footprint compared to $\pi_0$ model on the Orin platform, as summarized in Tab.~\ref{tab:model_comparison}. We collected 1K episodes of \texttt{table bussing} data using the G1 humanoid robot and fine-tuned both models. Despite having 12.5\% parameters, \textit{GigaBrain-0-Small} achieves a comparable success rate to $\pi_0$.
\begin{table}[htbp]
\centering
\caption{Model comparison between \textit{GigaBrain-0-Small} and $\pi_0$ on the Orin platform.}
\label{tab:model_comparison}
\begin{tabular}{lcccccc}
\toprule
Model & FLOPs & Parameters & Inference VRAM & Inference Latency & Success Rate \\
      & (GFLOPs) & (B/M) & (GB) & (seconds) & \\
\midrule
$\pi_0$ & 4400 & 3.2\,B & 17.5 & 1.28 & 80\% \\
\textit{GigaBrain-0-Small} & 840 & 402\,M & 1.9 & 0.13 &  80\% \\
\bottomrule
\end{tabular}
\end{table}

\section{Conclusion and Future Work}
\label{sec:conclusion}

In this work, we presented \textit{GigaBrain-0}, a vision-language-action model that leverages data generated by world models to overcome the scalability and diversity limitations of real-world robot data collection. By training on rich, photorealistic trajectories that span diverse scene appearances, object placements, and camera viewpoints, \textit{GigaBrain-0} achieves strong generalization across a wide spectrum of real-world robotic tasks, from dexterous manipulation to long-horizon mobile operations. Key architectural innovations, including RGBD input modeling and embodied Chain-of-Thought supervision, further enhance its spatial reasoning and sequential decision-making capabilities. Moreover, we introduced \textit{GigaBrain-0-Small}, a lightweight variant optimized for edge deployment on platforms such as the NVIDIA Jetson AGX Orin, demonstrating that VLA model can be made practical for real-time, on-device robotic control.

Looking ahead, our work opens several promising directions for future research. First, while we currently employ world models as scalable \textit{data engines}, a natural next step is to integrate them as interactive \textit{policy environments} for reinforcement learning. By enabling VLA agents to roll out trajectories and receive reward signals within the world model, we could drastically reduce the need for real-world trial-and-error while supporting policy refinement through simulated experience. Second, world models may learn universal representations of physical dynamics and task structure. Such representations could allow world models to evolve beyond passive simulators into active \textit{policy generators}, capable of proposing feasible action sequences or subgoals directly. Finally, closing the loop between VLA policies and world models through self-improvement cycles—where real-world rollouts continuously refine the world model, which in turn generates better training data, could pave the way toward truly autonomous, lifelong-learning robotic systems.



\clearpage
\setcitestyle{numbers}
\bibliographystyle{plainnat}
\bibliography{main}

\end{document}